\documentclass[10pt,twocolumn,letterpaper]{article}

\usepackage[pagenumbers]{cvpr}
\usepackage{multirow}
\usepackage{colortbl}      
\usepackage{xcolor}
\usepackage[accsupp]{axessibility}

\definecolor{cvprblue}{rgb}{0.21,0.49,0.74}
\usepackage[pagebackref,breaklinks,colorlinks,allcolors=cvprblue]{hyperref}

\makeatletter\renewcommand\paragraph{\@startsection{paragraph}{4}{\z@}
  {0.4em \@plus.0ex \@minus.2ex}{-.5em} {\normalfont\normalsize\bfseries}}\makeatother

\title{Image Diffusion Preview with Consistency Solver}

\author{
  Fu-Yun Wang$^{1,2}$\thanks{Work done while the author was a student researcher at Google DeepMind. Correspondence to Fu-Yun Wang (fywang0126@gmail.com) and Long Zhao (gary.zhao9012@gmail.com).} \quad
  Hao Zhou$^1$ \quad
  Liangzhe Yuan$^1$ \quad
  Sanghyun Woo$^1$ \quad
  Boqing Gong$^1$ \quad
  Bohyung Han$^{1,3}$ \quad \\
  Ming-Hsuan Yang$^1$ \quad
  Han Zhang$^1$ \quad
  Yukun Zhu$^1$ \quad
  Ting Liu$^1$ \quad
  Long Zhao$^1$ \\
  \\
  $^1$Google DeepMind \quad $^2$The Chinese University of Hong Kong  \quad $^3$Seoul National University\\
}

\begin{document}
\maketitle
\begin{abstract}
The slow inference process of image diffusion models significantly degrades interactive user experiences. To address this, we introduce Diffusion Preview, a novel paradigm employing rapid, low-step sampling to generate preliminary outputs for user evaluation, deferring full-step refinement until the preview is deemed satisfactory. Existing acceleration methods, including training-free solvers and post-training distillation, struggle to deliver high-quality previews or ensure consistency between previews and final outputs. We propose ConsistencySolver derived from general linear multistep methods, a lightweight,  trainable high-order solver optimized via Reinforcement Learning, that enhances preview quality and consistency. 
Experimental results demonstrate that ConsistencySolver significantly improves generation quality and consistency in low-step scenarios, making it ideal for efficient preview-and-refine workflows.
Notably, it achieves FID scores on-par with Multistep DPM-Solver using 47\% fewer steps, while outperforming distillation baselines. Furthermore, user studies indicate our approach reduces overall user interaction time by nearly 50\% while maintaining generation quality. 
Code is available at \url{https://github.com/G-U-N/consolver}.
\end{abstract}

\begin{figure}[t]
    \centering
    \includegraphics[width=0.9\linewidth]{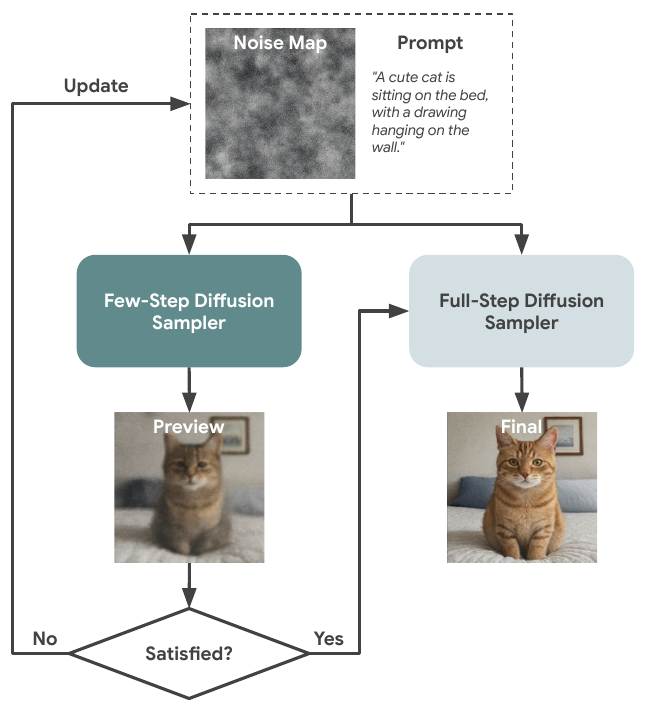}
    \vspace{-3mm}
    \caption{Overview of our \emph{Diffusion Preview} framework for efficient image generation using diffusion models.
Given a text prompt and a noise map, we first perform faster diffusion sampling to quickly generate a preview image. The user then decides whether the result is satisfactory. If not, they may refine the prompt or change the random seed. Once satisfied, full-step diffusion sampling is applied to generate the final high-quality image. This iterative workflow improves sampling efficiency and reduces unnecessary computational cost. }
    \label{fig:preview}
    \vspace{-3mm}
\end{figure}
\section{Introduction}
\label{sec:intro}

Diffusion models~\cite{ho2020ddpm} have significantly advanced generative artificial intelligence, particularly in high-fidelity visual data synthesis~\cite{diffusionbeatgan,rombach2022high,li2024autoregressive} and multimodal content creation~\cite{fan2025unified,podell2023sdxl}. Their ability to generate diverse, high-quality outputs has driven progress in various generative tasks. However, their computationally intensive inference process, requiring numerically solving the reverse differential equations, limits their practicality in resource-constrained settings~(\eg, mobile devices). To tackle this issue, we propose a \emph{preview-and-refine} framework, namely \emph{Diffusion Preview}, illustrated in \cref{fig:preview}, which splits the user's generation trials into two stages: (\textrm{i}) a rapid preview stage for generating and evaluating preliminary outputs and (\textrm{ii}) a refinement stage for resource-intensive high-quality sampling. Specifically, in the \emph{preview stage}, a fast, low-step sampling process generates a preliminary output that closely approximates the final high-quality result. This enables users to iterate quickly, experimenting with prompts or random seeds with minimal computational cost. In the \emph{refine stage}, when a preview meets expectations, the same iterated parameters will be used in a full-step sampling process to produce a high-fidelity output, fully leveraging the model’s capabilities.

This workflow is particularly valuable in interactive settings, such as design prototyping, where rapid feedback is critical. For instance, a designer can preview multiple image variations in seconds, select a promising candidate, and refine it into a polished result, saving significant time and resources.  
We  argue that a robust \emph{Diffusion Preview} framework should exhibit the following characteristics:
\begin{itemize}
    \item \textbf{Fidelity.} Previews should closely resemble the final output in visual and structural quality, providing reliable representations that enable informed user decisions while maintaining sufficient quality for effective evaluation.
    \item \textbf{Efficiency.} To support rapid iteration, the preview stage should minimize computational overhead, enabling users to quickly generate and explore multiple variations.
    \item \textbf{Consistency.} Previews should ensure a predictable and stable mapping between initial parameters (\eg, random seeds) and the final output, guaranteeing that refining a satisfactory preview produces a high-quality result aligned with user expectations.
\end{itemize}

We consider the diffusion sampling process based on the Probability Flow ODE~(PF-ODE) of diffusion models, as PF-ODE is a deterministic sampling algorithm~\cite{song2021sde}. When all initial parameters are fixed (e.g., prompts, initial noise), executing the exact PF-ODE sampling yields consistent results. This distinguishes PF-ODE from general SDE algorithms, as the sampling process does not introduce any additional random noise. We treat the exact PF-ODE sampling (termed full-step sampling) as the target for our refined results, aiming to achieve accurate previews of the final target through low-step sampling.

However, achieving effective \emph{Diffusion Preview} poses significant challenges for existing diffusion acceleration techniques. Training-free methods, such as zero-shot ODE solvers~\cite{lu2022dpm,lu2022dpmpp,song2021ddim,liu2022pseudo,karras2022edm}, rely on theoretical assumptions that may not align with the model's actual behavior. It frequently produces low-quality previews that fail to capture the essential characteristics of the final output.
Post-training approaches present different limitations. ODE distillation methods~\cite{luo2023latentconsistencymodelssynthesizing,song2023consistency} and score distillation techniques~\cite{dmdv2,diffinstruct,sid} bake acceleration directly into model weights, enabling high-quality outputs in a few steps but at substantial cost. These methods require expensive retraining and often disrupt the deterministic correspondence between noise space and data space induced by the PF-ODE. Moreover, ODE distillation methods suffer from accumulated distillation errors, causing degradation of the original ODE path and deterioration in generation quality. Score distillation methods fundamentally alter the model's learned trajectory due to their GAN-like training objectives~\cite{heusel2017gans,dmdv2}. Furthermore, distilled models typically lose key properties of the original diffusion models, such as flexible inference step selection and score estimation.

To this end, we introduce \emph{ConsistencySolver}, a novel solution tailored for the \emph{Diffusion Preview} paradigm. \emph{ConsistencySolver} is a trainable, high-order solver that optimizes the sampling dynamics of pre-trained diffusion models using \emph{Reinforcement Learning}~(RL)~\cite{sutton1998reinforcement}. By adapting to the model's sampling dynamics rather than modifying the model itself, \emph{ConsistencySolver} produces high-quality previews in low-step regimes while preserving the deterministic PF-ODE mapping essential for consistent refinement. \emph{ConsistencySolver} synergizes the strengths of efficient ODE solving and distillation learning, learning an improved sampling strategy directly from data while maintaining the base model's integrity and flexibility. 

In summary, our \textbf{main contributions} are: (\textrm{i}) A flexible, trainable solver framework that improves preview fidelity in low-step sampling scenarios; (\textrm{ii}) An RL-based optimization strategy for diffusion model sampling dynamics, offering a robust alternative to existing acceleration techniques; (\textrm{iii}) Comprehensive empirical experiments demonstrating that \emph{ConsistencySolver} achieves a superior balance among preview fidelity, efficiency, and consistency, enabling seamless \emph{Diffusion Preview} workflows.

\section{Related works}\label{sec:related}

Despite the superior generative quality of diffusion models since their inception~\cite{ho2020ddpm,song2019ncsn}, sampling latency remains a critical bottleneck relative to alternatives such as GANs~\cite{goodfellow2014generative} and VAEs~\cite{kingma2013auto}. 

\paragraph{Training-free ODE solvers.} Training-free acceleration hinges on optimized ODE solvers for the probability-flow ODE (PF-ODE)~\cite{song2021sde}. Early strides reduced NFE from 1000 to under 50 via deterministic~\cite{nichol2021improved} or quadratic timestep schedules~\cite{song2021ddim}, with Analytic-DPM~\cite{bao2022analytic} deriving closed-form optimal variance. Leveraging PF-ODE’s semi-linear structure, subsequent solvers approximate analytic integrals: DPM-Solver~\cite{lu2022dpm} employs Taylor expansion, DEIS~\cite{zhang2023deis} polynomial extrapolation, and iPNDM lower-order multistep warm-starts. Extensions include DPM-Solver++~\cite{lu2022dpmpp} (single- and multi-step variants), EDM~\cite{karras2022edm} (Heun’s method), PDNM~\cite{liu2022pseudo} (linear multistep with Runge-Kutta initialization), and UniPC~\cite{zhao2023unipc} (unified predictor-corrector), collectively pushing NFE toward 10.

\paragraph{Distilling ODE sampling dynamics.}  Distillation-based solvers, by contrast, train auxiliary networks to emulate multi-step trajectories in single-step predictions. Representative approaches encompass reparameterized DDPMs with KID loss~\cite{watson2021learning}, higher-order gradient prediction via truncated Taylor terms (GENIE~\cite{dockhorn2022genie}), intermediate timestep regression (AMED-Solver~\cite{zhou2024fastodebasedsamplingdiffusion}), and stepwise residual coefficients (D-ODE~\cite{kim2024distillingodesolversdiffusion}). Although differing in formulation, these methods converge on segment-wise trajectory matching (\ie, supervising single-step high-order inference with multi-step outputs), which yields locally consistent but globally suboptimal alignment. In opposition, our proposed framework introduces a generalized functional form, empirically validated via reinforcement learning to achieve superior efficiency, efficacy, and final-sample consistency.

\begin{figure*}[t]
    \centering
    \includegraphics[width=0.92\linewidth]{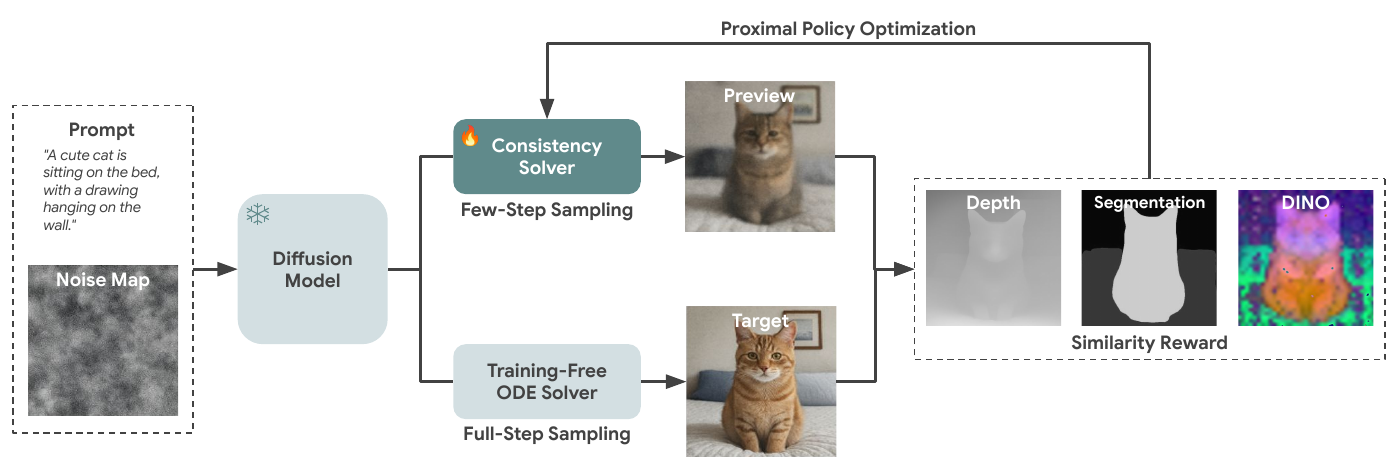}
    \vspace{-3mm}
    \caption{Overview of our RL framework for optimizing a learnable ODE solver in diffusion sampling. Given a prompt and a noise map, the diffusion model $\boldsymbol \epsilon_{\boldsymbol{\phi}}$ predicts denoising directions conditioned on the prompt. A learnable ODE solver $\mathbf \Psi_{\boldsymbol \theta}$ generates a preview image $\mathbf x_{\text {p}}$ via few-step sampling, while a training-free solver $\mathbf \Psi$ produces a target image $\mathbf x_{\text{gt}}$ using full-step sampling. The similarity reward $\mathcal{R}$ based on depth maps, segmentation masks, DINO features \etc guides the update of $\boldsymbol \theta$ via Proximal Policy Optimization (PPO).}
    \label{fig:method}
    \vspace{-3mm}
\end{figure*}

\section{Preliminaries on ODE solvers~\label{sec:preliminaries}}

Diffusion models~\cite{ho2020ddpm} generate samples by numerically integrating PF-ODE~\cite{song2021sde}. We start by reviewing the mathematical foundations of the PF-ODE and common solver approximations, and then discuss general linear multistep methods that leverage multiple prior states to improve convergence and accuracy.

\subsection{PF-ODE} Diffusion models define a series of intermediate distributions $\mathbb{P}_t(\mathbf{x} | \mathbf{x}_0) = \mathcal{N}(\alpha_t \mathbf{x}_0, \sigma_t^2 \mathbf{I})$, where $\mathbf{x}_0$ is the data. The noise adding process is formulated as the Stochastic Differential Equation (SDE)~\cite{song2019ncsn,song2021sde}:
$\mathrm{d}\mathbf{x}_t = f_t \mathbf{x}_t \mathrm{d}t + g_t \mathrm{d}\mathbf{w}_t$,
where $\mathrm d\mathbf{w}_t$ denotes the Wiener process, and the functions $f_t$ and $g_t$ are defined as: $
\mathrm df_t = \frac{\mathrm{d} \log \alpha_t}{\mathrm{d}t}$, $g_t^2 = \frac{\mathrm{d} \sigma_t^2}{\mathrm{d}t} - 2 \frac{\mathrm{d} \log \alpha_t}{\mathrm{d}t} \sigma_t^2\, . $
The deterministic reversal of the SDE~(\ie, PF-ODE) is given by~\cite{song2021sde}:
\begin{equation}~\label{eq:pf-ode-origin}
\mathrm{d} \mathbf{x}_t = \left[f_t\mathbf{x}_t - \frac{g^2_t}{2}\nabla_{\mathbf{x}_t}\log \mathbb{P}_t (\mathbf{x}_t)\right] \mathrm{d}t\, .
\end{equation}

\noindent Adopting $\boldsymbol \epsilon (\mathbf x_t, t) = -\sigma_t \nabla_{\mathbf x_t} \log \mathbb P_{t}(\mathbf x_t, t)
$, we can re-write \cref{eq:pf-ode-origin} into a simplified form: 
\begin{equation}\label{eq:pf-ode}
    \mathrm d  \left(\frac{\mathbf x_t}{\mathbf \alpha_t} \right) = \mathrm d \left(\frac{\sigma_t }{\alpha_t} \right) \cdot \boldsymbol \epsilon(\mathbf x_t, t)  \, .
\end{equation}

\subsection{Diffusion ODE solvers}
Denote $\mathbf y_t = \frac{\mathbf x_t}{\alpha_t}$, $\mathbf y_s = \frac{\mathbf x_s}{\alpha_s}$, $n_t = \frac{\sigma_t}{\alpha_t}$, $n_s = \frac{\sigma_s}{\alpha_s}$ in \cref{eq:pf-ode}, we can give the exact solution of the above PF-ODE:
\begin{equation}\label{eq:exact-solution}
    \begin{split}
    \mathbf y_s &= \mathbf y_t + \int_{n_t}^{n_s} \boldsymbol \epsilon (\mathbf x_{t_n}, t_n) \mathrm d n \, ,
    \end{split} 
\end{equation}
where $t_n$ is the inverse function of $n_t$. The key to obtaining the exact solution for \cref{eq:exact-solution} lies in how we approximate the integration from \( n_t \) to \( n_s \). Common techniques include: (i) \textit{naive approximation}, where assuming constant $\boldsymbol{\epsilon}(\mathbf{x}_t, t)$ over $[s,t]$ yields $\mathbf{y}_s = \mathbf{y}_t + (n_s - n_t) \boldsymbol{\epsilon}(\mathbf{x}_t, t)$, equivalent to DDIM~\cite{song2021ddim}; (ii) \textit{middle point approximation}, where a midpoint $r$ with $n_r = \sqrt{n_t \cdot n_s}$ gives $\mathbf{y}_s = \mathbf{y}_t + (n_s - n_t) \boldsymbol{\epsilon}(\mathbf{x}_r, r)$, equivalent to DPM-Solver-2~\cite{lu2022dpm}. These approximations can also be derived via Taylor expansion analysis (see the supplementary material).

\subsection{Linear multistep method} In addition to the above naive approximations, Linear Multistep Methods (LMMs)~\cite{sauer2018numerical,butcher2016numerical,hairer1993solving} are known to be effective for solving ODEs by utilizing multiple prior states to improve accuracy and speed up the convergence. Given an ODE of the form \( \frac{\mathrm d\mathbf{x}_t}{\mathrm dt} = f(\mathbf{x}_t, t) \), an \( m \)-step LMM approximates the solution \( \mathbf{x}_{t_{i+1}} \) using the recurrence:
\begin{multline}\label{eq:lmm}
    \mathbf{x}_{t_{i+1}} = \sum_{j=0}^{m-1} \mu_j \mathbf{x}_{t_{i-j}} + \\ (t_{i+1} - t_{i}) \sum_{j=0}^{m} w_j f(t_{i+1-j}, \mathbf{x}_{t_{i+1-j}})\, ,
\end{multline}
for \( i = m-1, m, \dots, N-1 \), where \( \mathbf{x}_{t_{i}}, \mathbf{x}_{t_{i-1}}, \dots, \mathbf{x}_{t_{i - m + 1}} \) are the \emph{state vectors} stored for the last $m$ steps, \( f \) represents the ODE’s derivative function, and \( \mu_j \) and \( w_j \) are approach-specific coefficients.   The method is \textit{explicit} if \( w_0 = 0 \), using only past states for the update, or \textit{implicit} if \( w_0 \neq 0 \), requiring a nonlinear solve at each step. Typically, explicit methods are favored for computational efficiency, while implicit methods enhance stability for stiff ODEs.

\section{ConsistencySolver\label{sec:method}}
\subsection{Adaptive ODE solvers for faithful previews}

To achieve high-fidelity, consistent previews in few-step diffusion sampling, we introduce \emph{ConsistencySolver}—a learnable, multistep ODE solver that dynamically adapts its integration strategy to maximize alignment between low-step previews and high-step reference generations. Unlike fixed solvers that apply rigid numerical schemes across all timesteps, \emph{ConsistencySolver} treats the choice of integration coefficients as a \emph{policy} to be optimized, conditioned on the local dynamics of the sampling trajectory.

Given a pretrained diffusion model $\boldsymbol{\epsilon}_{\boldsymbol{\phi}}(\mathbf{x}_t, t, \boldsymbol{c})$ where $\mathbf{x}_t$ is the noisy input at time $t$, and $\boldsymbol{c}$ is the conditioning signal (\eg, text prompt), we perform $N$-step sampling over discretized timesteps $\{t_i\}_{i=0}^N \subset [0,1]$. For clarity, we denote $\boldsymbol{\epsilon}_i \triangleq \boldsymbol{\epsilon}_{\boldsymbol{\phi}}(\mathbf{x}_{t_i}, t_i, \boldsymbol{c})$. At each transition from $t_i$ to $t_{i+1}$, \emph{ConsistencySolver} computes the update via a \textbf{weighted combination of past noise predictions}, followed by a deterministic ODE step. Specifically, $\mathbf {\Psi}_{\boldsymbol \theta}$ is formulated as:
\begin{equation}\label{eq:consistencysolver}
\mathbf{y}_{t_{i+1}} = \mathbf{y}_{t_i} + (n_{t_{i+1}} - n_{t_i}) \cdot  \left [ \sum_{j=1}^{m} w_j(t_i, t_{i+1}) \cdot \boldsymbol{\epsilon}_{i+1-j} \right ]\, ,
\end{equation}
where $\mathbf y_{t_i} = \frac{\mathbf x_{t_i}}{\alpha_{t_i}}$, $\mathbf x_{t_{i+1}}$ can be obtained by $\alpha_{t_{i+1}} \cdot \mathbf y_{t_{i+1}}$ , $n_t = \sigma_t / \alpha_t$, $m$ is the solver order (number of historical steps used), and the adaptive coefficients $w_j(t_i, t_{i+1})$ are generated by a lightweight neural policy network:
\begin{equation}
\begin{bmatrix} w_1 & w_2 & \cdots & w_{m} \end{bmatrix}^\top= \boldsymbol{f}_{\boldsymbol{\theta}}(t_i, t_{i+1})\, .
\end{equation}
The network $\boldsymbol{f}_{\boldsymbol{\theta}}$, which is implemented as an MLP with inputs $(t_i, t_{i+1})$, learns to predict context-aware integration weights that best preserve semantic and structural fidelity across step budgets. We provide a diagram illustrating the workflow of the generalized learnable ODE solver $\mathbf{\Psi}_{\boldsymbol{\theta}}$ in the supplementary material.

\paragraph{Training objective.} The training objective is to maximize preview–target consistency.  To be specific, let $\mathbf{x}_{\text{gt}}$ be the output of full-step sampling from initial noise $\mathbf{z} \sim \mathcal{N}(0,\mathbf{I})$ under prompt $\boldsymbol{c}$; let $\mathbf{x}_{\text{p}}$ be the output of few-step sampling using \emph{ConsistencySolver} with the same $\mathbf{z}$ and $\boldsymbol{c}$. Our goal is to find the optimal solver policy that achieves the highest similarity reward $\mathcal R = \mathrm{Sim}(\mathbf x_{\text{gt}}, \mathbf x_\text{p})$:
\begin{equation}
\mathbf{\Psi}_{\theta^*} = \arg\max_{\mathbf{\Psi}_{\boldsymbol{\theta}}} \mathbb{E}_{\mathbf{z},\boldsymbol{c}} \left[ \text{Sim}(\mathbf{x}_{\text{gt}}, \mathbf{x}_{\text{p}}) \right]\, ,
\end{equation}
where $\text{Sim}(\cdot,\cdot)$ is a perceptual similarity metric (\eg, depth maps, segmentation masks, DINO, \etc). 
This objective directly incentivizes the solver to produce previews that serve as reliable proxies for the final generation.

\paragraph{Solver searching via RL.} To discover an optimal adaptive multistep ODE solver, we cast the training of the policy network \(\boldsymbol{f}_{\boldsymbol{\theta}}\) as a sequential decision-making problem and optimize it with Proximal Policy Optimization (PPO)~\cite{schulman2017proximal}.

\textit{Offline dataset preparation.} Prior to training, we generate an offline dataset consisting of prompt–noise–reference triples \(\{(\boldsymbol{c}^{(k)}, \mathbf{z}^{(k)}, \mathbf{x}_{\text{gt}}^{(k)})\}_{k=1}^M\). For each entry, \(\boldsymbol{c}^{(k)}\) is sampled from the training prompt distribution, \(\mathbf{z}^{(k)} \sim \mathcal{N}(0,\mathbf{I})\), and \(\mathbf{x}_{\text{gt}}^{(k)}\) is generated via full-step sampling using the pretrained diffusion model. This dataset is fixed and reused across all experiments, enabling reproducible reward computation and eliminating the overhead of on-the-fly reference target generation during policy optimization.

\textit{Training episode rollout.} At each PPO episode, we uniformly sample a batch of \(B\) triples from the offline dataset. For each selected \((\boldsymbol{c}, \mathbf{z}, \mathbf{x}_{\text{gt}})\), we unroll a \(K\)-step preview trajectory using \(\mathbf{\Psi}_{\boldsymbol{\theta}}\) of~\cref{eq:consistencysolver}. At every transition (\(t_i \to t_{i+1}\)) within a predefined \(K\)-step schedule \(\{t_0 > t_1 > \cdots > t_K\}\), the policy processes inputs \((t_i, t_{i+1})\) through a lightweight MLP to output the coefficients sampling \(\boldsymbol{w}(t_i,t_{i+1}) = [w_1,\dots,w_{m}]\) and corresponding probabilities.

\textit{Reward and policy update.} Upon completing the \(K\)-step rollout, the preview \(\mathbf{x}_{\text{p}}\) is compared against the precomputed \(\mathbf{x}_{\text{gt}}\), yielding a scalar similarity reward \(\mathcal{R} = \text{Sim}(\mathbf{x}_{\text{gt}}, \mathbf{x}_{\text{p}})\). The policy is optimized via the standard PPO clipped surrogate objective:
\begin{equation}
\mathcal{J}_{\text{PPO}} = \mathbb{E} \left[ \min\!\bigl( r(\theta)\hat{A},\ \operatorname{clip}(r(\theta),1-\epsilon,1+\epsilon)\hat{A} \bigr) \right]\, ,
\end{equation}
where \(\theta\) denotes policy parameters, \(r(\theta) = \frac{\pi_{\theta}(a|s)}{\pi_{\theta_{\text{old}}}(a|s)}\) is the probability ratio between current and old policies, \(\hat{A}\) is the estimated advantage, \(\epsilon \in (0,1)\) is the clipping parameter, and \(\operatorname{clip}(\cdot, 1-\epsilon, 1+\epsilon)\) restricts \(r(\theta)\) to \([1-\epsilon, 1+\epsilon]\) to ensure stable updates. The advantage is computed with batch self-normalization:
\begin{equation}
\hat{A} = \frac{\mathcal{R} - \mathbb{E}[\mathcal{R}]}{\sigma[\mathcal{R}] + \delta}\, ,
\end{equation}
with \(\mathbb{E}[\mathcal{R}]\) and \(\sigma[\mathcal{R}]\) being the mean and standard deviation of rewards in the current minibatch, and \(\delta > 0\) a small constant to prevent division by zero. This follows common RL practice in generative modeling~\cite{li2023remax,shao2024deepseekmath,ahmadian2024back,black2023training,fan2024reinforcement}.

\subsection{Theoretical grounding}
While \emph{ConsistencySolver} is trained end-to-end via RL, its architectural form is rigorously derived from classical LMMs~\cite{sauer2018numerical,butcher2016numerical,hairer1993solving}, adapted to PF-ODEs.
Recall the general $m$-step LMM for  $\frac{\mathrm d\mathbf{x}_t}{\mathrm dt} = f(t, \mathbf{x}_t)$ in \cref{eq:lmm}. We adapt LMMs to PF-ODE sampling through three principled modifications:

\begin{enumerate}
    \item \textbf{Explicit-only design: $w_0 = 0$}. Empirical analyses show that PF-ODE trajectories are smooth and non-stiff~\cite{zhou2024fastodebasedsamplingdiffusion,chen2024trajectory}. Implicit solves are unnecessary and computationally prohibitive. Therefore, we only consider the explicit design by setting $w_0 = 0$.

    \item \textbf{Anchor to current state: $\mu_0 = 1$, $\mu_j = 0$ for $j \geq 1$}. We retain only the most recent state $\mathbf{y}_{t_i}$ as the integration base, eliminating redundant history storage while preserving high-order accuracy via derivative blending.
    
    \item \textbf{Timestep-conditioned coefficients}. Classical LMMs use fixed $w_j$ in \cref{eq:lmm}. We relax this to $w_j(t_i, t_{i+1})$, allowing the solver to adapt its integration paradigm as the denoising timesteps. 
    
\end{enumerate}
Notably, rather than deriving the coefficients in \cref{eq:consistencysolver} through theoretical assumptions or approximations, we treat them as learnable unknowns, which endows the \emph{ConsistencySolver} with exceptional flexibility and broad applicability.
We further demonstrate that several widely used diffusion solvers~\cite{song2021ddim,liu2022pseudo,lu2022dpm,lu2022dpmpp} can be recast within the \emph{ConsistencySolver} framework defined in \cref{eq:consistencysolver}. See the supplementary material for additional details.

\subsection{RL \vs distillation}
\emph{ConsistencySolver} is flexible in training, supporting either RL or distillation.
We choose to use RL due to its three key advantages compared with distillation methods: (\textrm{i}) \textit{Compatibility with non-differentiable rewards.} RL eliminates the need for a differentiable reward and avoids backpropagating through the diffusion trajectory, thereby removing a primary cause of instability and overhead in distillation.
(\textrm{ii}) \textit{Superior generalization and quality.} The RL-trained \emph{ConsistencySolver} better generalizes to novel prompt-noise pairs, yielding higher fidelity and elevated average consistency scores across CLIP, DINO, Depth and additional metrics (see \cref{tab:comparison}).
(\textrm{iii}) \textit{Reduced training overhead.} Relying solely on sparse rewards from the final clean output, RL forgoes intermediate gradient storage. Furthermore, only the compact MLP participates in loss computation, substantially lowering memory usage and facilitating efficient training.
In \cref{sec:quantitative}, we compare the proposed RL based \emph{ConsistencySolver} with distillation baselines (AMED \cite{zhou2024fastodebasedsamplingdiffusion} and  \textit{Ours-Distill}).
The experimental results empirically demonstrate the advantages of the proposed RL based method to distillation methods.

\section{Experiments}\label{sec:experiments}

\subsection{Experimental setup}\label{sec:setup}

We evaluate \emph{ConsistencySolver} using Stable Diffusion~\cite{rombach2022high} for text-to-image generation and FLUX.1-Kontext~\cite{labs2025flux} for instructional image editing. For each model, we sample 2,000 caption-noise-sample pairs from evaluation datasets, with ``ground truth'' samples ($\mathbf x_\text{gt}$) obtained using a 40-step multistep DPM-Solver. 
Unless otherwise specified, we use depth maps as the reward function in RL.
To evaluate \emph{Diffusion Preview}, we assess three core aspects: \textbf{fidelity}, \textbf{efficiency}, and \textbf{consistency}. These metrics ensure previews are accurate, efficient, and well-aligned with refined outputs, meeting the demands of high-quality image generation.

For text-to-image generation, the fidelity is measured using the Fréchet Inception Distance~(FID)~\cite{heusel2017gans}, which compares feature distributions between generated previews and real images. For instructional image editing, we adopt Edit Reward~\cite{wu2025editreward} and Edit Score~\cite{luo2025editscore} to measure the editing fidelity and the instruction alignment. The efficiency is quantified as inference time per image, reflecting the efficiency of preview generation. \cref{tab:consistency} summarizes the six dimensions we utilized for measuring consistency.

\begin{table*}[h]
\centering
\caption{Metrics employed for consistency evaluation.}
\vspace{-2mm}
\resizebox{0.8\linewidth}{!}{
\begin{tabular}{@{}llcc@{}}
\toprule
\textbf{Dimension} & \textbf{Description} & \textbf{Model} & \textbf{Metric} \\ \midrule
Semantic alignment~(CLIP) & Image semantic measured by vision embeddings & CLIP ViT-L/14~\cite{radford2021learning} & Cosine similarity \\
Structural consistency~(DINO) & Alignment in image structure and layout & DINOv2-L/14~\cite{oquab2023dinov2} & Cosine similarity \\
Perceptual similarity~(Inc.) & Visual resemblance through perception models & Inception V3~\cite{szegedy2016rethinking} & Cosine similarity \\
Segmentation accuracy~(Seg.) &  Overlaps between segmentation masks & SegFormer~\cite{segformer} & Mean Dice coefficient \\
Pixel-level similarity~(Img.) & Pixel-wise differences between raw images & -- & PSNR \\
Depth consistency~(Dep.) & Differences between depth maps & Depth Anything V2~\cite{yang2024depth} & PSNR \\ \bottomrule
\end{tabular}}
\label{tab:consistency}
\vspace{-3mm}
\end{table*}

\paragraph{Evaluation datasets.} For text-to-image generation with Stable Diffusion, we use the prompts from the validation set of COCO 2017~\cite{lin2014microsoft} as the prompts for evaluation, which is a common dataset adopted to assess the generation capacity of text-to-image diffusion models. For instructional image editing, we use KontextBench~\cite{labs2025flux} as the reference images and editing instructions to reflect the model's performance regarding aspects such as character reference, global editing, local editing, \etc.

\paragraph{Distillation baselines.} We use trajectory based distillation methods as our distillation baselines. Two methods are selected: AMED~\cite{zhou2024fastodebasedsamplingdiffusion} and \textit{Ours-Distill}. \textit{Ours-Distill} distills the full sampling trajectory by aligning intermediate states in a segment-wise fashion, sharing similar principles with AMED~\cite{zhou2024fastodebasedsamplingdiffusion} and D-ODE~\cite{kim2024distillingodesolversdiffusion}. More details are discussed in the supplementary material.

\begin{table}[t]
\centering
\caption{Comparison of \emph{ConsistencySolver} with baselines at various steps. Best results per step in \textbf{bold}. Ours-Distill is the proposed \emph{ConsistencySolver} with coefficients trained with trajectory distillation. AMED is only applicable to even steps. }
\label{tab:comparison}
\resizebox{\linewidth}{!}{
\begin{tabular}{@{}l|c|c|ccccccc@{}}
\toprule
\textbf{Method} & \textbf{Steps} & \textbf{FID}$\downarrow$ & \textbf{CLIP}$\uparrow$ & \textbf{Seg.}$\uparrow$ & \textbf{Dep.}$\uparrow$ & \textbf{Inc.}$\uparrow$ & \textbf{Img.}$\uparrow$ & \textbf{DINO}$\uparrow$ \\
\hline
\multicolumn{9}{c}{Training-Free ODE Solvers} \\
\hline
DDIM~\cite{song2021ddim} & 5 & 52.59 & 87.8 & 41.9 & 14.2 & 74.1 & 16.4 & 73.2 \\
iPNDM~\cite{liu2022pseudo} & 5 & 37.44 & 89.3 & 44.1 & 14.3 & 75.7 & 15.0 & 73.9 \\
\rowcolor{black!5}UniPC~\cite{zhao2023unipc} & 5 & 23.15 & 93.2 & 67.2 & 18.7 & 85.0 & 19.6 & 85.5 \\
\rowcolor{black!5}DEIS~\cite{zhang2023deis} & 5 & 25.78 & 92.2 & 65.4 & 18.4 & 83.8 & 19.2 & 84.3 \\
\rowcolor{black!5}Multistep DPM~\cite{lu2022dpm} & 5 & 25.87 & 93.1 & 66.6 & 19.1 & 85.6 & 20.6 & 85.5 \\
\hline
DDIM & 8 & 29.46 & 91.1 & 54.2 & 16.2 & 81.5 & 17.9 & 79.9 \\
iPNDM & 8 & 25.88 & 91.7 & 54.9 & 16.4 & 81.7 & 17.0 & 79.8 \\
\rowcolor{black!5}UniPC & 8 & 19.68 & 95.5 & 75.0 & 21.4 & 90.0 & 21.6 & 90.5 \\
\rowcolor{black!5}DEIS & 8 & 20.14 & 94.9 & 73.6 & 20.7 & 89.1 & 21.0 & 89.3 \\
\rowcolor{black!5}Multistep DPM & 8 & 19.53 & 95.9 & 76.3 & 21.8 & 90.8 & 23.2 & 90.6 \\
\hline
DDIM & 10 & 24.88 & 92.4 & 59.0 & 17.1 & 83.6 & 18.7 & 82.1 \\
iPNDM & 10 & 22.65 & 92.9 & 59.1 & 17.4 & 84.0 & 18.0 & 82.6 \\
\rowcolor{black!5}UniPC & 10 & 19.38 & 96.5 & 79.7 & 23.2 & 91.9 & 23.0 & 92.5 \\
\rowcolor{black!5}DEIS & 10 & 19.42 & 95.9 & 77.2 & 21.9 & 90.8 & 22.1 & 91.4 \\
\rowcolor{black!5}Multistep DPM & 10 & 19.29 & 97.0 & 80.5 & 24.1 & 93.1 & 25.1 & 93.0 \\
\hline
DDIM & 12 & 22.81 & 93.1 & 61.6 & 17.7 & 85.1 & 19.2 & 83.4 \\
iPNDM & 12 & 21.23 & 93.5 & 62.7 & 17.9 & 85.3 & 18.6 & 83.9 \\
\rowcolor{black!5}UniPC & 12 & 19.32 & 97.2 & 82.5 & 24.7 & 93.4 & 24.4 & 94.1 \\
\rowcolor{black!5}DEIS & 12 & 19.24 & 96.6 & 80.2 & 23.2 & 92.3 & 23.2 & 92.9 \\
\rowcolor{black!5}Multistep DPM & 12 & 18.95 & 97.7 & 84.4 & 25.9 & 94.7 & 26.8 & 94.5 \\
\hline
\multicolumn{9}{c}{Distillation-Based Methods} \\
\hline
DMD2~\cite{dmdv2} & 1 & 19.88 & 89.3 & 42.1 & 12.6 & 70.5 & 12.1 & 73.8 \\
LCM~\cite{luo2023latentconsistencymodelssynthesizing} & 2 & 22.20 & 89.8 & 51.9 & 14.9 & 77.9 & 14.3 & 75.4 \\
LCM & 4 & 22.00 & 90.0 & 50.8 & 14.3 & 78.1 & 12.6 & 75.1 \\
LCM & 8 & 21.50 & 89.2 & 47.3 & 13.6 & 76.5 & 10.7 & 72.9 \\
PCM~\cite{wang2024phased} & 4 & 21.67 & 92.6 & 63.0 & 17.2 & 83.9 & 17.9 & 82.3 \\
Rectified Diff.~\cite{wangrectified} & 4 & 20.64 & 94.4 & 67.6 & 18.5 & 87.0 & 19.7 & 85.6 \\
\hline
\multicolumn{9}{c}{Distillation-Based Solvers} \\
\hline
\rowcolor{black!5}AMED~\cite{zhou2024fastodebasedsamplingdiffusion} & 4 & 31.09 & 90.4 & 58.6 & 16.9 & 80.4 & 17.9 & 80.8 \\
\rowcolor{black!5}AMED & 6 & 20.42 & 93.3 & 66.2 & 18.1 & 85.5 & 18.8 & 85.4 \\
\rowcolor{black!5}AMED & 8 & 19.22 & 94.9 & 72.4 & 20.0 & 88.3 & 20.5 & 88.8 \\
\rowcolor{black!5}AMED & 10 & 18.95 & 96.2 & 78.3 & 22.2 & 91.4 & 22.3 & 91.8 \\
\rowcolor{black!5}AMED & 14 & 19.08 & 97.2 & 83.3 & 24.4 & 93.6 & 24.3 & 94.3 \\
\rowcolor{black!5}Ours-Distill & 5 & 22.91 & 92.9 & 66.7 & 18.5 & 84.8 & 19.3 & 85.1 \\
\rowcolor{black!5}Ours-Distill & 8 & 19.65 & 95.1 & 74.0 & 20.8 & 89.3 & 21.1 & 89.5 \\
\rowcolor{black!5}Ours-Distill & 10 & 19.29 & 95.9 & 77.5 & 22.0 & 91.0 & 22.2 & 91.5 \\
\rowcolor{black!5}Ours-Distill & 12 & 19.06 & 96.6 & 80.4 & 23.0 & 92.3 & 23.2 & 92.8 \\
\hline
\multicolumn{9}{c}{Proposed Method} \\
\hline
\rowcolor{black!5}ConsistencySolver & 5 & \textbf{20.39} & \textbf{94.2} & \textbf{69.4} & \textbf{19.3} & \textbf{87.1} & \textbf{20.8} & \textbf{86.5} \\
\rowcolor{black!5}ConsistencySolver & 8 & \textbf{18.82} & \textbf{96.4} & \textbf{78.5} & \textbf{22.2} & \textbf{91.6} & \textbf{23.4} & \textbf{91.2} \\
\rowcolor{black!5}ConsistencySolver & 10 & \textbf{18.66} & \textbf{97.2} & \textbf{83.2} & \textbf{24.9} & \textbf{93.9} & \textbf{25.3} & \textbf{93.5} \\
\rowcolor{black!5}ConsistencySolver & 12 & \textbf{18.53} & \textbf{97.9} & \textbf{85.6} & \textbf{26.7} & \textbf{95.1} & \textbf{26.7} & \textbf{95.0} \\
\bottomrule
\end{tabular}
}
\vspace{-3mm}
\end{table}

\subsection{Quantitative comparison}\label{sec:quantitative}

\paragraph{Stable Diffusion.}
\cref{tab:comparison} presents a comprehensive quantitative comparison of \emph{ConsistencySolver} against various baselines on Stable Diffusion for text-to-image generation across multiple measures including FID and consistency metrics.
Among training-free ODE solvers such as DDIM, iPNDM, and multistep DPM-Solver, \emph{ConsistencySolver} consistently outperforms at equivalent step counts.
It achieves lower FID values (\eg, $20.39$ at 5 steps \vs multistep DPM-Solver's $25.87$) and higher consistency scores across all dimensions, demonstrating superior alignment with refined outputs.
Compared with distillation-based methods such as DMD2, Rectified Diffusion, LCM, and PCM, which often require fewer steps but sacrifice quality, \emph{ConsistencySolver} delivers competitive or better performance.
For instance, at 4 to 8 steps, it surpasses LCM and PCM in FID and most consistency metrics, highlighting its efficiency in balancing speed and quality without distillation overhead.
As the number of steps increases (\eg, up to 12), \emph{ConsistencySolver} further refines its outputs, yielding the best overall results with FID as low as $18.53$ and peak consistency scores like $97.9$ in CLIP and $95.1$ in Inception.

\paragraph{FLUX.1-Kontext.} In \cref{tab:kontext_comparison}, we compare \emph{ConsistencySolver} with baselines in terms of Edit Reward (E.\ R.) and Edit Score (E.\ S.) for fidelity and instruction alignment, alongside consistency metrics (DINO, Inception, CLIP, and Depth). At lower steps (3 to 4), \emph{ConsistencySolver} shows marked improvements over FLUX.1-Kontext, with higher Edit Reward (\eg, $0.73$ at 4 steps \vs $0.61$) and Edit Score ($5.67$ \vs $5.45$), indicating better editing accuracy and adherence to instructions. By 5 steps, it achieves the best results across all metrics, including a superior Edit Reward of $0.86$ and Depth consistency of $25.18$, underscoring its ability to produce high-fidelity previews that closely match refined edits while maintaining computational efficiency.

\begin{table}[t]
\centering
\caption{Comparison of \emph{ConsistencySolver} with FLUX.1-Kontext at various steps. Best results per step in \textbf{bold}.}
\label{tab:kontext_comparison}
\vspace{-1mm}
\resizebox{\linewidth}{!}{
\begin{tabular}{@{}l|c|cc|cccc@{}}
\toprule
\textbf{Method} & \textbf{Steps} & \textbf{E.\ R.}$\uparrow$ & \textbf{E.\ S.}$\uparrow$ & \textbf{DINO}$\uparrow$ & \textbf{Inc.}$\uparrow$ & \textbf{CLIP}$\uparrow$ & \textbf{Dep.}$\uparrow$ \\
\hline
\multirow{3}{*}{Euler~\cite{karras2022edm}} 
  & 3 & 0.32 & 4.66 & 88.47 & 83.56 & 92.68 & 22.90 \\
  & 4 & 0.61 & 5.45 & 91.31 & 86.75 & 93.95 & 23.99 \\
  & 5 & 0.79 & 5.80 & 93.09 & 89.16 & 95.25 & 24.76 \\
\hline
\multirow{3}{*}{Heun~\cite{karras2022edm}} 
  & 3 & -0.23 & 3.08 & 82.13 & 77.28 & 89.70 & 20.25 \\
  & 4 & -0.40 & 2.21 & 77.83 & 72.39 & 86.95 & 17.85 \\
  & 5 & 0.18 & 4.31 & 87.69 & 81.73 & 92.28 & 21.71 \\
\hline
\multirow{3}{*}{DPM} 
  & 3 & 0.39 & 4.90 & 89.36 & 84.68 & 93.14 & 23.20 \\
  & 4 & 0.69 & 5.60 & 92.10 & 88.22 & 94.64 & 24.30 \\
  & 5 & 0.80 & 5.88 & 93.68 & 90.33 & 95.71 & 25.05 \\
\hline
\multirow{3}{*}{Multistep DPM} 
  & 3 & 0.41 & 5.04 & 89.20 & 84.87 & 93.11 & 22.96 \\
  & 4 & 0.72 & 5.57 & 91.83 & 88.12 & 94.49 & 23.70 \\
  & 5 & 0.83 & 5.92 & 93.44 & 90.17 & 95.53 & 24.59 \\
\hline
\multirow{3}{*}{ConsistencySolver} 
  & 3 & \textbf{0.45} & \textbf{5.13} & \textbf{89.38} & \textbf{85.01} & \textbf{93.10} & \textbf{23.08} \\
  & 4 & \textbf{0.73} & \textbf{5.67} & \textbf{92.39} & \textbf{88.71} & \textbf{94.86} & \textbf{24.27} \\
  & 5 & \textbf{0.86} & \textbf{6.02} & \textbf{93.90} & \textbf{90.76} & \textbf{95.87} & \textbf{25.18} \\
\bottomrule
\end{tabular}
\vspace{-5mm}
}
\end{table}

\paragraph{Cross-model generalization.} To evaluate the transferability of \emph{ConsistencySolver}, we directly apply the solver trained on SD1.5 to unseen models with different weights, architectures, and scales, including SD1.4, DreamShaper, and SDXL, without any retraining. As shown in \cref{tab:sdxl_sd14_comparison}, \emph{ConsistencySolver} consistently outperforms all training-free ODE solvers on unseen models. On SDXL with 10 steps, it achieves an FID of $23.32$, substantially outperforming Multistep DPM-Solver ($26.32$) and DEIS ($26.40$). On SD1.4 with 5 steps, the improvement is even more significant, with an FID of $20.22$ compared to $25.22$ for Multistep DPM-Solver. Qualitative comparisons are provided in the supplementary material. These results suggest that different diffusion models share similar optimal sampling dynamics.

\begin{table}[t]
\centering
\caption{Cross-model generalization on COCO 2017. \emph{ConsistencySolver} trained on SD1.5 is directly applied to unseen models without retraining. Best results in \textbf{bold}.}
\label{tab:sdxl_sd14_comparison}
\vspace{-1mm}
\resizebox{0.92\linewidth}{!}{
\begin{tabular}{@{}l|c|c|cc@{}}
\toprule
\textbf{Model} & \textbf{Steps} & \textbf{Method} & \textbf{FID}$\downarrow$ & \textbf{CLIP}$\uparrow$ \\
\midrule
\multirow{5}{*}{SDXL} & \multirow{5}{*}{10}
  & DDIM & 27.80 & 32.23 \\
& & iPNDM & 30.07 & 31.68 \\
& & \cellcolor{black!5}DEIS & \cellcolor{black!5}26.40 & \cellcolor{black!5}32.51 \\
& & \cellcolor{black!5}Multistep DPM & \cellcolor{black!5}26.32 & \cellcolor{black!5}32.52 \\
\cmidrule(l){3-5}
& & \cellcolor{black!5}\emph{ConsistencySolver} (Ours) & \cellcolor{black!5}\textbf{23.32} & \cellcolor{black!5}\textbf{33.45} \\
\midrule
\multirow{5}{*}{SD1.4} & \multirow{5}{*}{5}
  & DDIM & 32.50 & 29.37 \\
& & iPNDM & 36.16 & 27.20 \\
& & \cellcolor{black!5}DEIS & \cellcolor{black!5}25.83 & \cellcolor{black!5}29.90 \\
& & \cellcolor{black!5}Multistep DPM & \cellcolor{black!5}25.22 & \cellcolor{black!5}29.94 \\
\cmidrule(l){3-5}
& & \cellcolor{black!5}\emph{ConsistencySolver} (Ours) & \cellcolor{black!5}\textbf{20.22} & \cellcolor{black!5}\textbf{30.16} \\
\bottomrule
\end{tabular}
}
\vspace{-3mm}
\end{table}

\subsection{Qualitative comparison}\label{sec:qualitative}
\begin{figure*}[t]
    \centering
\includegraphics[width=0.94\linewidth]{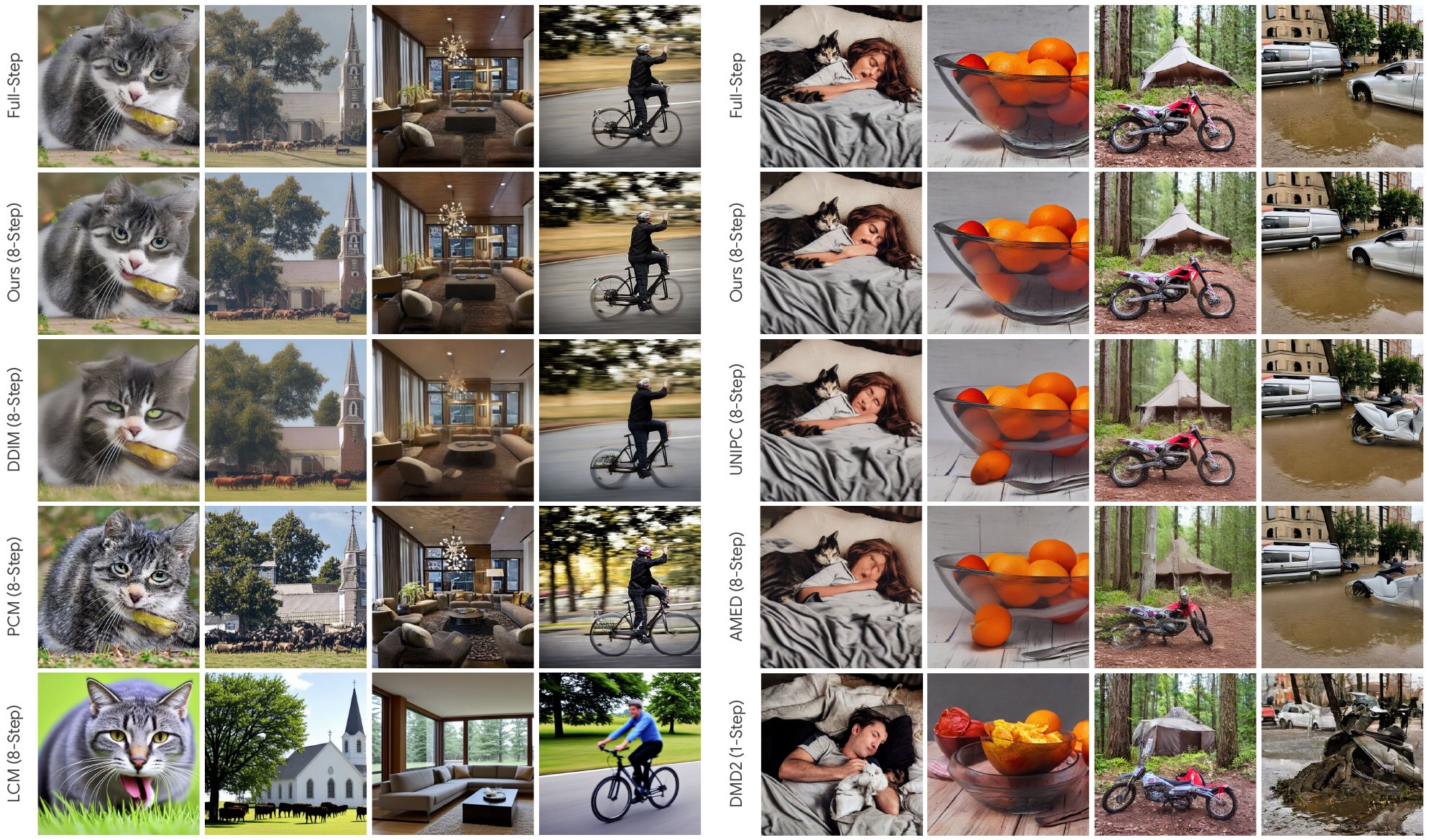}
\vspace{-3mm}
    \caption{Visual comparison on Stable Diffusion for text-to-image generation.}
    \label{fig:visualization}
    \vspace{-3mm}
\end{figure*}

\begin{figure}[t]
    \centering
\includegraphics[width=0.92\columnwidth]{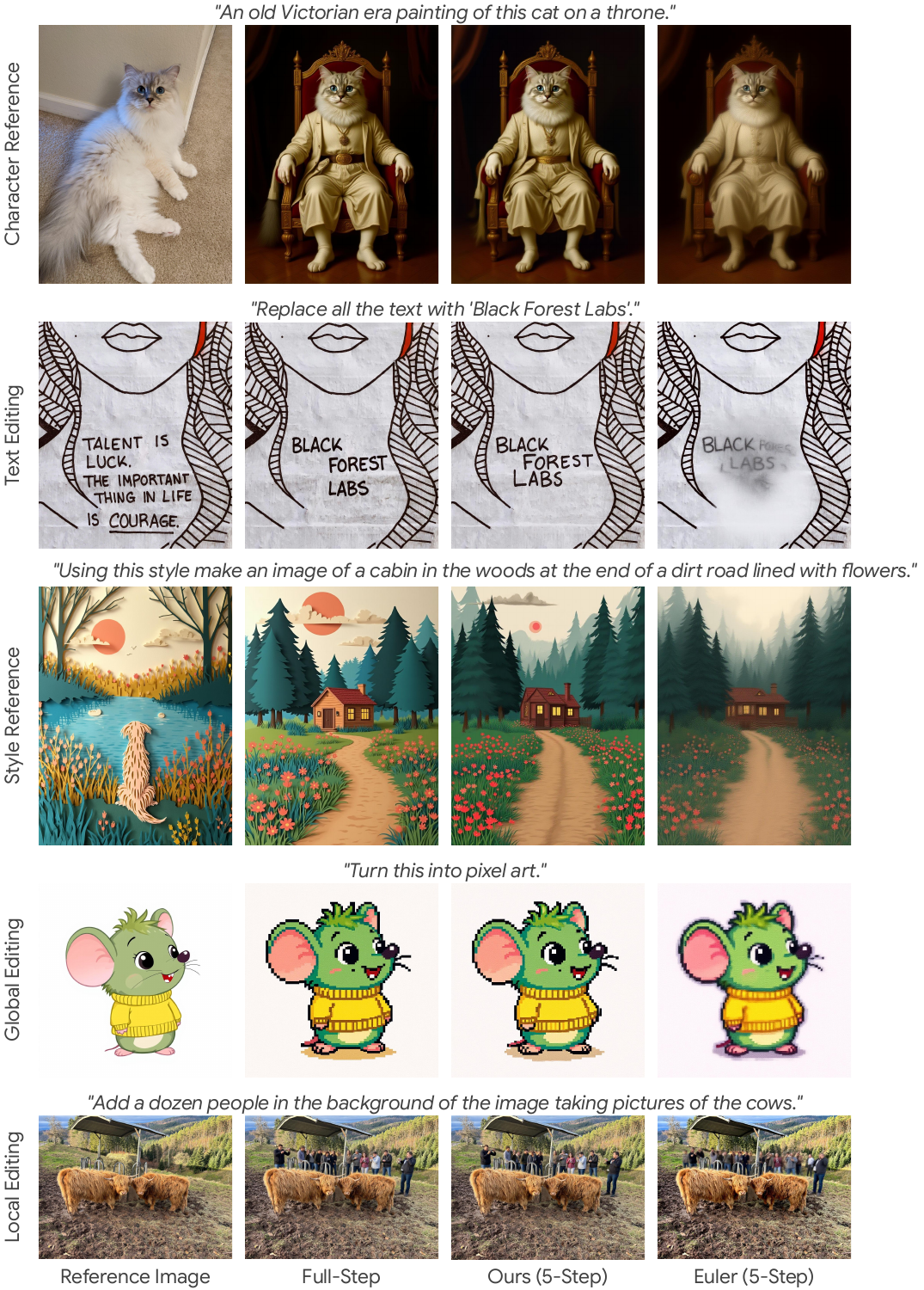}
\vspace{-1mm}
    \caption{Visual comparison on FLUX.1-Kontext for instructional image editing. Previews are generated with 5 inference steps.}
    \label{fig:edit_visualization}
\vspace{-3mm}
\end{figure}

\cref{fig:visualization} presents visual comparisons of previews generated by Stable Diffusion for text-to-image tasks, while 
\cref{fig:edit_visualization} shows visual comparisons of previews produced by FLUX.1-Kontext for instructional image editing. We demonstrate five representative examples across diverse editing tasks, including character reference, text editing, style reference, global editing, and local editing.  Compared to training-free ODE solvers and distillation-based methods, \emph{ConsistencySolver} yields previews with sharper details and superior alignment to the refined outputs.

\subsection{Studies on \emph{Diffusion Preview}}
In addition to the aforementioned evaluations on generation quality and consistency, we further validate the practical effectiveness of our proposed preview-and-refine paradigm through user study.
Specifically, we fix the prompt and repeatedly sample images with different random noise until the users are satisfied or the attempt limit is reached.
We then compare the average time and attempts used by different methods to generate the user satisfactory images.
Besides real human users, we also use Claude Sonnet 4 as a proxy evaluator to avoid potential bias in human judgment.
To demonstrate the efficiency gains of our preview mechanism, we conduct comparisons with two modes.

In the \textbf{full mode}, for a given prompt, we generate the image using a 40-step multistep DPM-Solver. The output is evaluated using both Claude Sonnet 4 and human judgment to determine whether it meets expectations.

In the \textbf{preview mode}, we first generate a fast preview using an 8-step \emph{ConsistencySolver} and assess it via the same judgment mechanism. If the preview fails to meet requirements, a new preview is generated; otherwise, we perform one 40-step DPM-Solver refinement (\ie, full-step sampling is triggered only after confirming a satisfied preview).

We report the average end-to-end inference time (including denoising and VAE decoding) for both paradigms. To prevent cases where Stable Diffusion fundamentally fails to satisfy certain prompts from skewing the results, we impose a maximum of 10 attempts per prompt. Prompts that remain unsatisfactory after 10 trials are discarded, ensuring that timing statistics accurately reflect the efficiency of the preview mechanism under normal conditions.

To evaluate generalizability across diverse user needs, we use three validation prompt sets: GenEval prompts~\cite{geneval}, COCO 2017 validation~\cite{lin2014microsoft}, and LAION~\cite{laion}. Detailed experimental protocols, including LLM prompts and human evaluation guidelines, are provided in the supplementary material. As shown in \cref{tab:preview_efficiency}, \emph{Diffusion Preview} reduces average inference time by up to $55\%$ on LAION with only a minor increase in attempts~(\ie, $6.00$ $\rightarrow$ $6.35$). %

\paragraph{Comparison to distillation.}  
As distillation-based models continue to improve, particularly the emergening of state-of-the-art single-step models like DMD2~\cite{dmdv2}, a natural question arises: \emph{do we still need the preview-and-refine paradigm?}  If the generation quality is sufficiently high, one might argue that the \textit{Diffusion Preview} paradigm and consistency property become less critical.  

To investigate this, we use Claude Sonnet 4 to conduct a user-centric evaluation.
We record the number of prompts satisfied within 10 attempts.
As shown in \cref{tab:consistency_matters}, though DMD2 achieves competitive FID scores, it satisfies significantly fewer prompts compared to both the base model and our \emph{ConsistencySolver}.
On the GenEval prompts, DMD2 with and without GAN satisfy only $57.0\%$ and $47.1\%$ of the prompts compared with the base model, while our method maintains $94.2\%$.
This disparity reveals a critical insight: \emph{despite the competitive FID scores achieved by distillation-based methods, the loss of consistency fundamentally undermines generation quality in ways not captured by distribution-level metrics}. For the proposed preview-and-refine workflows, where users rely on previews to guide iterative refinement, maintaining consistency is essential.

\begin{table}[t]
\centering
\caption{End-to-end inference time (seconds) and average attempts on H100 across three prompt sets.}
\label{tab:preview_efficiency}
\vspace{-1mm}
\resizebox{\linewidth}{!}{
\begin{tabular}{@{}lcc cc cc c@{}}
\toprule
\multirow{2.4}{*}{\textbf{Evaluator}} &
\multicolumn{2}{c}{\textbf{GenEval}} &
\multicolumn{2}{c}{\textbf{COCO 2017}} &
\multicolumn{2}{c}{\textbf{LAION}} &
\multirow{2.4}{*}{\textbf{Speedup}}\\
\cmidrule(lr){2-3} \cmidrule(lr){4-5} \cmidrule(lr){6-7}
& Full & Preview & Full & Preview & Full & Preview & \\
\midrule
\multicolumn{8}{c}{\textbf{Inference time (seconds)}} \\
\midrule
Claude Sonnet & 2.88 & 1.74 & 3.64 & 1.85 & 6.35 & 2.87 & 1.88$\times$ \\
Human & 3.82 & 2.16 & 3.52 & 2.03 & 5.18 & 2.58 & 1.85$\times$ \\
\midrule
\multicolumn{8}{c}{\textbf{Average attempts}} \\
\midrule
Claude Sonnet & 3.00 & 3.12 & 3.71 & 3.60 & 6.00 & 6.35 & -- \\
Human & 3.55 & 3.80 & 3.30 & 3.42 & 5.17 & 5.45 & -- \\
\bottomrule
\end{tabular}
}
\vspace{-3mm}
\end{table}

\begin{table}[t]
  \centering
  \caption{User satisfaction within 10 attempts. Despite competitive FID, distillation methods show significant satisfaction drops, highlighting the practical importance of consistency.}
  \label{tab:consistency_matters}
  \vspace{-1mm}
  \resizebox{0.96\linewidth}{!}{
    \begin{tabular}{@{}l cc cc@{}}
      \toprule
      \multirow{2.4}{*}{\textbf{Method}} &
      \multicolumn{2}{c}{\textbf{COCO 2017}} &
      \multicolumn{2}{c}{\textbf{GenEval}} \\
      \cmidrule(lr){2-3} \cmidrule(lr){4-5}
      & Satisfied & \% of Base & Satisfied & \% of Base \\
      \midrule
      Base model (40-step) & 2,143 & 100.0\% & 121 & 100.0\% \\
      \midrule
      DMD2 w/ GAN          & 1,389 & 64.8\%  & 69  & 57.0\% \\
      DMD2 w/o GAN         & 1,267 & 59.1\%  & 57  & 47.1\% \\
      \textbf{ConsistencySolver (8-step)} & \textbf{2,057} & \textbf{96.0\%} & \textbf{114} & \textbf{94.2\%} \\
      \bottomrule
    \end{tabular}
  }
\end{table}

\subsection{Ablation study}\label{sec:ablation}

\paragraph{Solver orders.}
We assess the effect of solver order $m$ in \cref{eq:consistencysolver} at 5, 8, and 10 steps. Order 4 consistently achieves the best overall performance, striking an effective balance between accuracy and RL search complexity. Lower orders show reduced structural fidelity, while Order 5 yields only marginal gains. We adopt Order 4 as the default. Detailed comparisons are provided in the supplementary material.

\paragraph{Reward models.}
We investigate the impact of different reward models on the RL training of \emph{ConsistencySolver}. Among the six reward choices (Depth, Inception, CLIP, Img., DINO, Seg.), the Depth reward provides the most balanced trade-off between structural consistency and overall robustness across all step counts. We therefore adopt Depth as the default reward. Detailed comparisons are provided in the supplementary material.

\section{Conclusion}
This paper proposes \textit{Diffusion Preview}, a novel paradigm aimed at generating fast and consistent approximations of diffusion model outputs to enable efficient previewing in generative modeling. To address this task, we introduce \emph{ConsistencySolver}, a method that delivers reliable previews with few steps, outperforming existing training-free and distillation-based approaches in consistency, paving the way for more practical generative modeling workflows.

\section*{Acknowledgments}
We thank Hartwig Adam and Florian Schroff from Google DeepMind for their support of this project, and Prof.\ Hongsheng Li from CUHK for his valuable advice.

{
    \small
    \bibliographystyle{ieeenat_fullname}
    \bibliography{refs}
}

  \clearpage
  \appendix
  \setcounter{page}{1}
  \setcounter{table}{0}
  \setcounter{figure}{0}
  \setcounter{equation}{0}
  \renewcommand{\thetable}{S\arabic{table}}
  \renewcommand{\thefigure}{S\arabic{figure}}
  \renewcommand{\theequation}{S\arabic{equation}}

  \makeatletter
  \def\maketitlesupplementary{
     \newpage
     \twocolumn[
        \centering
        \Large
        \textbf{\thetitle}\\
        \vspace{0.5em}Supplementary Material \\
        \vspace{1.0em}
        \normalsize
        Fu-Yun Wang$^{1,2}$, Hao Zhou$^1$, Liangzhe Yuan$^1$, Sanghyun Woo$^1$,
  Boqing Gong$^1$, Bohyung Han$^{1,3}$,\\
        Ming-Hsuan Yang$^1$, Han Zhang$^1$, Yukun Zhu$^1$, Ting Liu$^1$, Long
  Zhao$^1$\\
        \vspace{0.5em}
        $^1$Google DeepMind \quad $^2$The Chinese University of Hong Kong \quad
  $^3$Seoul National University\\
        \vspace{1.0em}
     ]
  }
  \makeatother

  \maketitlesupplementary
\vspace{1em}

\begin{figure*}[h]
    \centering
    \includegraphics[width=0.95\linewidth]{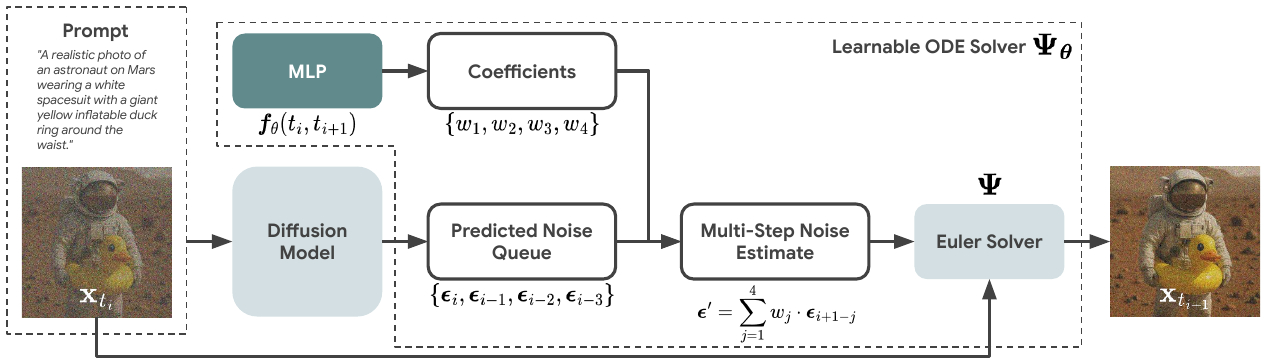}
    \caption{Workflow of the generalized learnable ODE solver $\mathbf{\Psi}_{\boldsymbol \theta}$ with Order 4~($m=4$). At each sampling step, the diffusion model predicts noise $\epsilon_i$ conditioned on the input prompt and timestep. A learnable neural network $\boldsymbol f_\theta$ generates adaptive coefficients ${w}_j$, $j=1,2,3,4$ from current timestep $t_i$, and target timestep $t_{i+1}$ ,  which are used to form a multi-step noise estimate $\boldsymbol \epsilon' = \sum_{j=1}^{4} w_j \cdot \boldsymbol \epsilon_{i+1-j}$. The ODE solver $\mathbf{\Psi}_{\boldsymbol \theta}$ then updates the sample from $\mathbf x_{t_i}$ to $\mathbf x_{t_{i+1}}$. This approach enables more accurate and stable integration in the generative sampling process.}
    \label{fig:mlp}
\end{figure*}

\section{Common diffusion ODE solvers via Taylor expansion}
\label{sec:taylor-expansion}

The exact solution of \cref{eq:exact-solution} requires numerical approximation of
\begin{equation}
\Delta \mathbf{y}_{t \to s} = \int_{n_t}^{n_s} \boldsymbol{\epsilon}(\mathbf{x}_{t_n}, t_n) \, \mathrm{d}n.
\end{equation}
Let $h = n_s - n_t$. The Taylor expansion of the integrand around $n_t$ yields
\begin{multline}
\label{eq:taylor-integral}
\int_{n_t}^{n_s} \boldsymbol{\epsilon}(\mathbf{x}_{t_n}, t_n) \, \mathrm{d}n 
= h \, \boldsymbol{\epsilon}(\mathbf{x}_t, t) 
  + \frac{h^2}{2} \, \frac{\mathrm{d}}{\mathrm{d}n} \boldsymbol{\epsilon}(\mathbf{x}_{t_n}, t_n) \Big|_{n_t}
   \\ + \frac{h^3}{6} \, \frac{\mathrm{d}^2}{\mathrm{d}n^2} \boldsymbol{\epsilon}(\mathbf{x}_{t_n}, t_n) \Big|_{n_t}
  + \cdots.
\end{multline}
  For brevity, we denote
\begin{equation}
\boldsymbol{\epsilon}_t \triangleq \boldsymbol{\epsilon}(\mathbf{x}_t, t), 
\end{equation}
and similarly for other time points~(\eg, $s$).

\subsection{First-order: DDIM / Euler (naïve)}
\begin{equation}
\Delta \mathbf{y}_{t \to s} \approx h \, \boldsymbol{\epsilon}_t \, .
\end{equation}
Retains only the zeroth-order term in \cref{eq:taylor-integral}.

\subsection{Second-order: DPM-Solver-2 / midpoint}
The midpoint method uses one evaluation near the interval center:
\begin{equation}
\label{eq:midpoint-method}
\Delta \mathbf{y}_{t \to s} \approx h \, \boldsymbol{\epsilon}_r, 
\quad n_r \approx n_t + \frac{h}{2}\, .
\end{equation}

To see second-order accuracy, approximate the missing derivative with a centered finite difference:
\begin{equation}
\frac{\mathrm{d}}{\mathrm{d}n} \boldsymbol{\epsilon} \Big|_{n_t}
\approx \frac{\boldsymbol{\epsilon}_r - \boldsymbol{\epsilon}_t}{h/2}.
\end{equation}
Insert into the desired second-order truncation:
\begin{align}
h \, \boldsymbol{\epsilon}_t 
+ \frac{h^2}{2} \cdot \frac{\boldsymbol{\epsilon}_r - \boldsymbol{\epsilon}_t}{h/2}
&= h \, \boldsymbol{\epsilon}_t 
 + h \, \bigl( \boldsymbol{\epsilon}_r - \boldsymbol{\epsilon}_t \bigr) \notag \\
&= h \, \boldsymbol{\epsilon}_r.
\end{align}
Thus $h \, \boldsymbol{\epsilon}_r$ exactly matches the second-order Taylor integral when the first derivative is estimated by a midpoint difference.  
DPM-Solver-2 exploits this insight, typically choosing $n_r = \sqrt{n_t n_s}$ (geometric midpoint in noise-scale space).

\section{Common diffusion ODE solvers interpreted using \emph{ConsistencySolver}}\label{sec:cases}
\emph{ConsistencySolver} treats the coefficients in \cref{eq:consistencysolver} as learnable unknowns. Here we show that several widely adopted diffusion solvers~\cite{song2021ddim,lu2022dpm,liu2022pseudo} can be easily interpreted using the form of \emph{ConsistencySolver}.

For notational simplicity, we denote $\boldsymbol \epsilon_{\boldsymbol \phi}(\mathbf x_{t_i}, t_i)$ simply as $\boldsymbol \epsilon_{i}$ throughout this section.

\textbf{DDIM (naive approximation)} performs the update:
    \begin{equation}
        \mathbf y_{t_{i+1}} = \mathbf y_{t_i} + (n_{t_{i+1}} - n_{t_i})\boldsymbol{\epsilon}_{i} \, .
    \end{equation}
    Comparing with \cref{eq:consistencysolver}, we can have the naive approximation corresponds to a one-step method ($m=1$) with the coefficient $w_1 = 1$.

\textbf{PNDM} utilizes the explicit 4-step Adams-Bashforth method~\cite{sauer2018numerical}. For the Ininial Value Problem~(IVP) $\mathrm d\mathbf y/ \mathrm dn = \boldsymbol{\epsilon}$, the update is:
    \begin{equation}
      \mathbf y_{t_{i+1}} = \mathbf y_{t_i} + \frac{\Delta n_i}{24} \left[ 55\boldsymbol{\epsilon}_i - 59\boldsymbol{\epsilon}_{i-1} + 37\boldsymbol{\epsilon}_{i-2} - 9\boldsymbol{\epsilon}_{i-3} \right] \, ,
    \end{equation}
    where $\Delta n_i = n_{t_{i+1}} - n_{t_i}$.
    This corresponds to $m=4$ with coefficients:
    \begin{equation}
        w_1 = \frac{55}{24}, \quad w_2 = -\frac{59}{24}, \quad w_3 = \frac{37}{24}, \quad w_4 = -\frac{9}{24} \, ,
    \end{equation}
    of the proposed the \emph{ConsistencySolver} defined in \cref{eq:consistencysolver}.

\textbf{DPM-Solver-2 (midpoint approximation)} uses an evaluation at an intermediate point $t_i$ (corresponding to $n_{t_i} = \sqrt{n_{t_{i-1}} n_{t_{i+1}}}$):
    \begin{equation}
    \begin{split}
    \mathbf y_{t_{i}} & = \mathbf y_{t_{i-1}} + (n_{t_{i}} - n_{t_{i-1}})\boldsymbol \epsilon_{i-1} \, , 
    \\
        \mathbf y_{t_{i+1}} & = \mathbf y_{t_{i-1}} + (n_{t_{i+1}} - n_{t_{i-1}})\boldsymbol{\epsilon}_{i} \\
        & = \mathbf y_{t_i} + (n_{t_{i+1}} - n_{t_{i-1}})\boldsymbol{\epsilon}_{i}   - (n_{t_{i}} - n_{t_{i-1}})\boldsymbol \epsilon_{i-1} \\
        & = \mathbf y_{t_i} + (n_{t_{i+1}} - n_{t_i})[ \frac{(n_{t_{i+1}} - n_{t_{i-1}})}{(n_{t_{i+1}} - n_{t_i})}\boldsymbol{\epsilon}_{i} \\ &  - \frac{(n_{t_{i}} - n_{t_{i-1}})}{(n_{t_{i+1}} - n_{t_i})}\boldsymbol \epsilon_{i-1}]
        \end{split}
    \end{equation}
        Comparing with \cref{eq:consistencysolver}, we can have DPM-Solver-2 corresponds to two-stages computation. When $i$ is even (\ie, $0,2,4,\dots$), the approximation corresponds to a one-step method ($m=1$) with the coefficient $w_1 = 1$. When $i$ is odd, the approximation corresponds to a two-step method ($m=2$) with the coefficient $w_1 = \frac{(n_{t_{i+1}} - n_{t_{i-1}})}{(n_{t_{i+1}} - n_{t_i})}, w_2= - \frac{(n_{t_{i}} - n_{t_{i-1}})}{(n_{t_{i+1}} - n_{t_i})}$.

\section{Visualization of \emph{ConsistencySolver}}
We visualize the computation paradigm of the proposed \emph{ConsistencySolver} in \cref{fig:mlp}, taking Order 4~($m=4$) as an example.

\section{Implementation details}

\subsection{ConsistencySolver training}

\paragraph{Training dataset.} We randomly sample 2,000 prompts from the LAION dataset~\cite{laion} and generate corresponding images using a 40-step multistep DPM-Solver, forming noise-prompt-target image triplets as our training data.

\paragraph{Training procedure.} All experiments are conducted on a single H100 GPU. For each training iteration, we select one prompt-noise pair and replicate it 80 times. We then apply the trainable \emph{ConsistencySolver} to generate 80 different sampling trajectories with random perturbations. Following the PPO algorithm, we increase the probability of high-reward trajectories while suppressing low-reward ones. By default, we use Order-4 solver configurations. The MLP network in \emph{ConsistencySolver} is trained from scratch using a learning rate of $1\times 10^{-4}$ for 3,000 iterations, requiring approximately 12 H100 GPU hours in total.

\subsection{Distillation baseline training}

Beyond the proposed RL-based training approach, we explore distillation-based alternatives to optimize the dynamic coefficients in \emph{ConsistencySolver}. We investigate two distillation schemes:

\paragraph{Final-state distillation.} This approach treats the entire few-step diffusion sampling chain as differentiable and directly uses the negative reward at the final state as the loss function. Gradients are backpropagated through the complete inference chain to optimize the parameters. While conceptually straightforward, this method exhibits significant drawbacks. First, backpropagating through the entire chain requires computing gradients not only for the \emph{ConsistencySolver} MLP but also for the underlying diffusion model (typically containing billions of parameters), substantially increasing computational cost. Second, we observe severe training instability, with the MLP failing to converge effectively in practice.

\paragraph{Trajectory distillation.} Inspired by prior work~\cite{zhou2024fastodebasedsamplingdiffusion,wang2024phased}, we propose a trajectory-based distillation method, referred to as \emph{Ours-Distill} in the main text. This approach requires storing the complete 40-step trajectory from the multistep DPM-Solver (introducing additional storage overhead). The objective is to match each intermediate state in the few-step \emph{ConsistencySolver} sampling to corresponding states in the 40-step reference trajectory. For example, when performing 5-step sampling, each \emph{ConsistencySolver} step should align with 8 steps of the reference solver. We use the negative similarity between these states as the loss function for backpropagation. This method significantly outperforms final-state distillation but still falls short of the RL-based approach, as demonstrated in our quantitative comparisons in \cref{tab:comparison}.

\paragraph{Training dataset.} We use the same 2,000 training samples as for \emph{ConsistencySolver} training to ensure fair comparison.

\subsection{Preview study experimental protocol}

\paragraph{Evaluation datasets.} For the preview study, we evaluate on three datasets: (1) GenEval evaluation set containing 553 prompts~\cite{geneval}, (2) COCO 2017 validation set with 5,000 prompts~\cite{lin2014microsoft}, and (3) 5,000 randomly sampled prompts from LAION~\cite{laion}.

\paragraph{Evaluation with LLM.} We use Claude Sonnet 4 as an automated judge to simulate a discerning user. The system prompt is designed to enforce strict evaluation criteria:

\begin{quote}
\textit{``You are a very picky user evaluating an AI-generated image for the prompt `\{prompt\}'. Be extremely critical---only approve if it perfectly matches the description in composition, quality, details, and realism. Respond with ONLY `SATISFIED' if it's perfect, or `NOT\_SATISFIED: [brief reason]' otherwise. Keep the reason under 50 words.''}
\end{quote}

This ensures the LLM judges each generated image with high standards, accepting only those that closely align with the prompt requirements.

\begin{table}[t]
\centering
\caption{Ablation study on model structure at 8 and 10 steps. Best results per metric in \textbf{bold}.}
\label{tab:ablation_structure}
\resizebox{0.99\linewidth}{!}{
\begin{tabular}{@{}l|c|ccccc@{}}
\toprule
\textbf{Model} & \textbf{Steps} & \textbf{Dep.}$\uparrow$ & \textbf{Inc.}$\uparrow$ & \textbf{Img.}$\uparrow$ & \textbf{CLIP}$\uparrow$ & \textbf{DINO}$\uparrow$ \\
\hline
\multicolumn{7}{c}{8 Steps} \\
\hline
Hidden Dim 32 & 8 & 22.08 & 91.55 & 23.07 & 96.24 & 90.96 \\
Hidden Dim 256 & 8 & \textbf{22.22} & \textbf{91.68} & \textbf{23.56} & \textbf{96.36} & \textbf{91.14} \\
Hidden Dim 1024 & 8 & 21.82 & 91.30 & 22.36 & 96.04 & 90.57 \\
Deep (12-Layer MLP) & 8 & 22.00 & 91.20 & 22.60 & 96.14 & 90.68 \\
\hline
\multicolumn{7}{c}{10 Steps} \\
\hline
Hidden Dim 32 & 10 & 24.68 & 93.67 & 24.80 & 97.16 & 93.31 \\
Hidden Dim 256 & 10 & \textbf{25.01} & \textbf{93.85} & \textbf{25.57} & \textbf{97.30} & \textbf{93.67} \\
Hidden Dim 1024 & 10 & 24.12 & 93.23 & 23.96 & 96.92 & 92.67 \\
Deep (12-Layer MLP) & 10 & 24.38 & 93.39 & 24.22 & 96.99 & 93.12 \\
\bottomrule
\end{tabular}
}
\end{table}
\paragraph{Human evaluation.} To complement LLM evaluation, we conduct human studies with real users. For each prompt, we pre-generate 10 images and record their generation times. These images are organized into questionnaires where participants sequentially evaluate whether each image satisfies the prompt. Participants stop at the first satisfactory image; if all images are unsatisfactory, the trial is discarded as discussed in the main text. We recruit 20 volunteers, each responsible for evaluating 100 prompts uniformly sampled across all test datasets, resulting in comprehensive human feedback on the practical effectiveness of our preview mechanism.

\subsection{Ablation study on model structures}
We analyze architectural variants of \emph{ConsistencySolver}, varying hidden dimension size and testing a deep 12-layer MLP with residual LayerNorm, evaluated at 8 and 10 steps. According to \cref{tab:ablation_structure}, the 256-dimensional model consistently outperforms others, delivering superior results in image similarity, semantic alignment, and overall consistency. Larger dimensions (\eg, 1024) slightly enhance depth estimation but compromise balance and efficiency. The deep MLP variant shows no meaningful advantage over the standard 256-dim architecture, suggesting that moderate capacity is sufficient for the task.

\subsection{Ablation study on solver orders}

We assess the effect of solver order, \ie, $m$ in \cref{eq:consistencysolver}, on \emph{ConsistencySolver}'s preview consistency at 5, 8, and 10 steps. As shown in \cref{tab:ablation_order}, Order 4 consistently achieves the best overall performance across step counts, leading in key structural and perceptual metrics while maintaining strong semantic alignment. Lower-order solvers (\eg, Order 2 or 3) show reduced fidelity in layout and depth consistency, whereas Order 5 yields only marginal improvements in minor dimensions likely due to the increased RL search space complexity. Overall, Order 4 strikes a better balance between efficiency and complexity.

\begin{table}[t]
\centering
\caption{Ablation study on solver order at 5, 8, and 10 steps. Best results per metric in \textbf{bold}.}
\vspace{-1mm}
\label{tab:ablation_order}
\resizebox{0.98\columnwidth}{!}{
\begin{tabular}{@{}l|c|cccccc@{}}
\toprule
\textbf{Orders} & \textbf{Steps} & \textbf{Dep.}$\uparrow$ & \textbf{Inc.}$\uparrow$ & \textbf{Seg.}$\uparrow$ & \textbf{Img.}$\uparrow$ & \textbf{CLIP}$\uparrow$ & \textbf{DINO}$\uparrow$ \\
\hline
\multicolumn{8}{c}{5 Steps} \\
\hline
Order 2 & 5 & \textbf{19.33} & \textbf{87.30} & 69.36 & \textbf{20.84} & 94.40 & 86.39 \\
Order 3 & 5 & 19.15 & 86.46 & 68.93 & 20.26 & 93.80 & 85.83 \\
Order 4 & 5 & 19.29 & 87.07 & \textbf{69.42} & 20.75 & 94.22 & 86.35 \\
Order 5 & 5 & \textbf{19.33} & 87.16 & 69.38 & 20.64 & \textbf{94.33} & \textbf{86.44} \\
\hline
\multicolumn{8}{c}{8 Steps} \\
\hline
Order 2 & 8 & 22.12 & 91.59 & 78.56 & 23.34 & 96.31 & 91.03 \\
Order 3 & 8 & 22.14 & 91.57 & 77.92 & 23.20 & 96.26 & 90.81 \\
Order 4 & 8 & \textbf{22.15} & \textbf{91.65} & \textbf{78.52} & \textbf{23.43} & \textbf{96.35} & \textbf{91.09} \\
Order 5 & 8 & 22.12 & 91.65 & 78.19 & 23.15 & 96.33 & 90.97 \\
\hline
\multicolumn{8}{c}{10 Steps} \\
\hline
Order 2 & 10 & 24.72 & 93.74 & 82.86 & 25.16 & \textbf{97.25} & 93.45 \\
Order 3 & 10 & 24.66 & 93.74 & 82.68 & 25.23 & 97.23 & 93.29 \\
Order 4 & 10 & \textbf{24.94} & \textbf{93.88} & \textbf{83.22} & \textbf{25.32} & \textbf{97.25} & \textbf{93.48} \\
Order 5 & 10 & 24.72 & 93.79 & 82.78 & 24.88 & 97.18 & 93.36 \\
\bottomrule
\end{tabular}
}
\end{table}

\subsection{Ablation study on reward models}

We investigate the impact of different reward models on the RL training of \emph{ConsistencySolver}. As shown in \cref{tab:ablation_reward}, the Depth reward provides strong structural fidelity, consistently achieving good performance across all steps. Meanwhile, the Img.\ reward performs well in pixel-level fidelity, particularly at higher steps. Although CLIP and DINO show competitive results in semantic alignment, Depth offers a more balanced trade-off between structural consistency and overall robustness. We therefore adopt Depth as the default reward for its reliable generalization across diverse evaluation scenarios.

\begin{table}[t]
\centering
\caption{Ablation study on reward model choice at 5, 8, and 10 steps. Best results per metric in \textbf{bold}.}
\vspace{-1mm}
\label{tab:ablation_reward}
\resizebox{0.98\columnwidth}{!}{
\begin{tabular}{@{}l|c|cccccc@{}}
\toprule
\textbf{Rewards} & \textbf{Steps} & \textbf{Dep.}$\uparrow$ & \textbf{Inc.}$\uparrow$ & \textbf{Seg.}$\uparrow$ & \textbf{Img.}$\uparrow$ & \textbf{CLIP}$\uparrow$ & \textbf{DINO}$\uparrow$ \\
\hline
\multicolumn{8}{c}{5 Steps} \\
\hline
Dep. & 5 & 19.29 & 87.07 & 69.42 & \textbf{20.75} & 94.22 & 86.35 \\
Inc. & 5 & 19.20 & 87.05 & 69.49 & 20.18 & 94.29 & 86.30 \\
CLIP & 5 & \textbf{19.32} & \textbf{87.30} & \textbf{69.73} & 20.30 & \textbf{94.46} & 86.50 \\
Img. & 5 & \textbf{19.32} & 87.22 & 69.44 & 20.69 & 94.40 & \textbf{86.53} \\
DINO & 5 & 19.29 & 87.19 & 69.64 & 20.43 & 94.39 & 86.43 \\
Seg. & 5 & 19.16 & 86.81 & 69.28 & 19.85 & 94.12 & 86.01 \\
\hline
\multicolumn{8}{c}{8 Steps} \\
\hline
Dep. & 8 & \textbf{22.15} & 91.65 & \textbf{78.52} & \textbf{23.43} & \textbf{96.35} & \textbf{91.09} \\
Inc. & 8 & 22.00 & 91.51 & 77.33 & 22.67 & 96.17 & 90.61 \\
CLIP & 8 & 21.94 & 91.45 & 77.54 & 22.56 & 96.15 & 90.75 \\
Img. & 8 & 22.11 & \textbf{91.75} & 78.17 & 23.39 & 96.34 & 90.97 \\
DINO & 8 & 22.03 & 91.62 & 77.84 & 22.99 & 96.28 & 90.87 \\
Seg. & 8 & 21.82 & 91.36 & 77.05 & 22.41 & 96.05 & 90.39 \\
\hline
\multicolumn{8}{c}{10 Steps} \\
\hline
Dep. & 10 & \textbf{24.94} & \textbf{93.88} & \textbf{83.22} & 25.32 & 97.25 & \textbf{93.48} \\
Inc. & 10 & 24.17 & 93.35 & 82.01 & 24.33 & 97.05 & 92.68 \\
CLIP & 10 & 24.25 & 93.44 & 81.84 & 24.14 & 96.99 & 92.76 \\
Img. & 10 & 24.80 & 93.87 & 82.74 & \textbf{25.37} & \textbf{97.28} & 93.39 \\
DINO & 10 & 24.49 & 93.60 & 82.55 & 24.81 & 97.15 & 93.01 \\
Seg. & 10 & 23.73 & 93.15 & 81.37 & 24.04 & 96.96 & 92.44 \\
\bottomrule
\end{tabular}
}
\end{table}

\section{Latency analysis across hardware configurations}
\label{sec:hardware-latency}

We present a detailed comparison of preview-and-refine latency by incorporating user reaction time across different hardware configurations. Following prior studies on human visual perception~\cite{amano2006estimation}, we assume a user reaction time of approximately 0.3 seconds per sample. \cref{tab:hardware_comparison} reports the end-to-end inference time on three datasets with different hardware. On high-performance GPUs (\eg, H100), the overall speedup is reduced when accounting for user interaction time. However, when deploying on M4 chips and CPUs (Platinum 8480), the sampling time per trial is substantially longer, making the human evaluation overhead negligible in comparison. These results highlight that the efficiency gains of \emph{Diffusion Preview} are particularly pronounced in resource-constrained deployment scenarios, where the preview-and-refine paradigm achieves up to $2.04\times$ speedup even after accounting for user interaction time.

\begin{table}[t]
\centering
\caption{End-to-end inference time (seconds) across hardware configurations.}
\label{tab:hardware_comparison}
\vspace{-3mm}
\resizebox{\linewidth}{!}{
\begin{tabular}{@{}lcc cc cc c@{}}
\toprule
\multirow{2.4}{*}{\textbf{Hardware}} &
\multicolumn{2}{c}{\textbf{GenEval}} &
\multicolumn{2}{c}{\textbf{COCO 2017}} &
\multicolumn{2}{c}{\textbf{LAION}} &
\multirow{2.4}{*}{\textbf{Speedup}}\\
\cmidrule(lr){2-3} \cmidrule(lr){4-5} \cmidrule(lr){6-7}
& Full & Preview & Full & Preview & Full & Preview & \\
\midrule
\multicolumn{8}{c}{\textbf{Without user interaction time}} \\
\midrule
SD1.5 (H100) & 3.82 & 2.16 & 3.52 & 2.03 & 5.18 & 2.58 & 1.84$\times$ \\
SD1.5 (M4) & 135.68 & 69.99 & 126.13 & 66.81 & 197.60 & 83.78 & 2.06$\times$ \\
SD1.5 (CPU) & 145.30 & 76.35 & 135.07 & 72.80 & 211.61 & 91.72 & 2.02$\times$ \\
\midrule
\multicolumn{8}{c}{\textbf{With user interaction time} (300\,ms per trial)} \\
\midrule
SD1.5 (H100) & 4.89 & 3.30 & 4.51 & 3.06 & 6.73 & 4.22 & 1.52$\times$ \\
SD1.5 (M4) & 136.75 & 71.13 & 127.12 & 67.84 & 199.15 & 85.42 & 2.04$\times$ \\
SD1.5 (CPU) & 146.37 & 77.49 & 136.06 & 73.83 & 213.16 & 93.36 & 2.00$\times$ \\
\bottomrule
\end{tabular}
}
\end{table}

\section{Cross-model generalization}
\label{sec:cross-model}

\begin{figure}[t]
    \centering
    \includegraphics[width=1.0\linewidth]{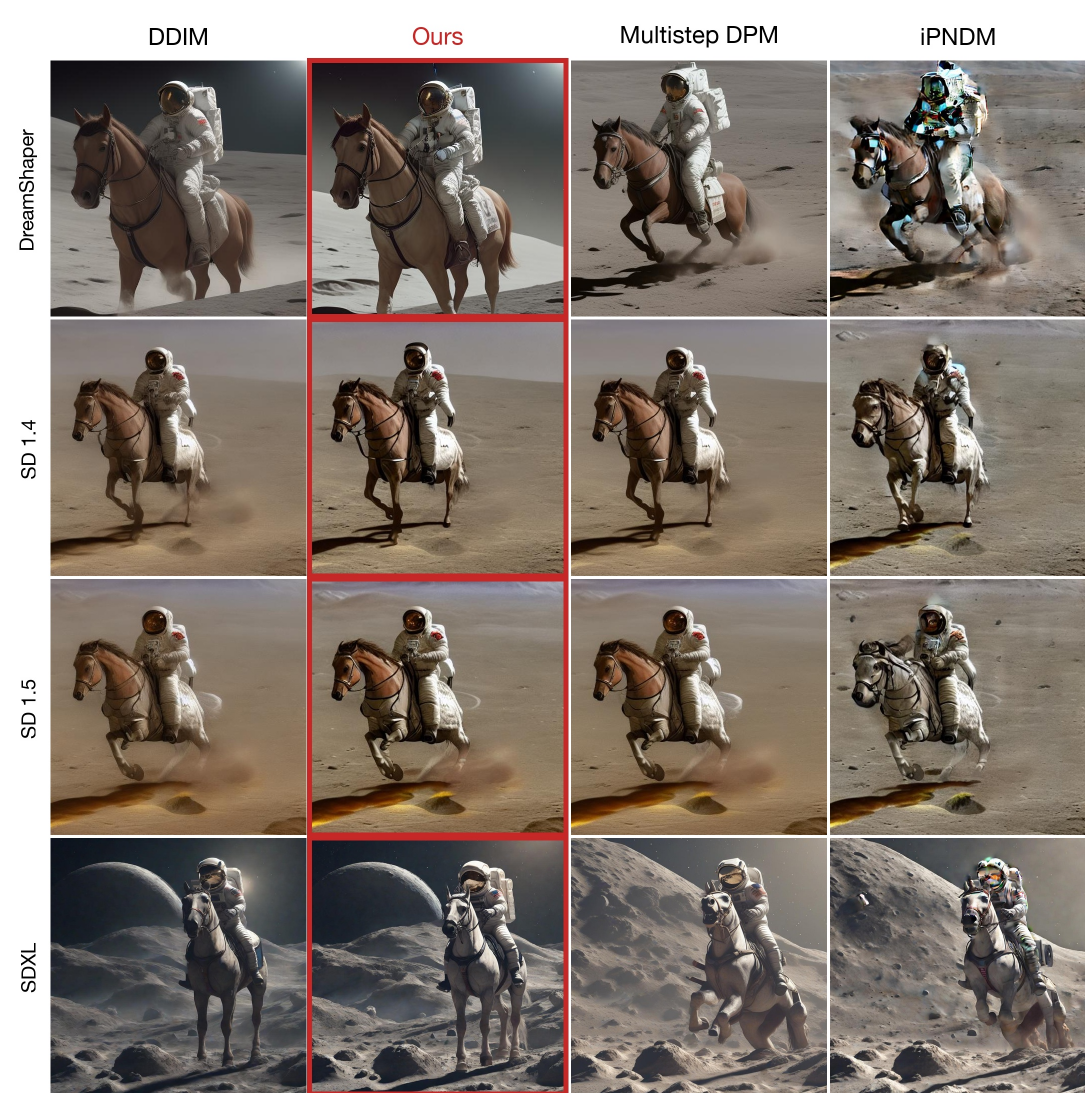}
    \vspace{-2mm}
    \caption{\emph{ConsistencySolver} trained on SD1.5 generalizes effectively to SD1.4, DreamShaper, and SDXL, achieving superior visual quality compared to training-free ODE solvers.}
    \label{fig:cross_model_demos}
\end{figure}

We provide qualitative comparisons for the cross-model generalization experiments discussed in the main text (\cref{tab:sdxl_sd14_comparison}). \cref{fig:cross_model_demos} shows that \emph{ConsistencySolver} trained on SD1.5 produces visually superior results on SD1.4, DreamShaper, and SDXL compared to training-free ODE solvers, further confirming its strong transferability across model families.

\end{document}